\documentclass[preprint,12pt]{elsarticle}

\usepackage{lineno,hyperref}
\modulolinenumbers[5]
\usepackage{multirow}

\journal{International Journal of Forecasting} 

\usepackage[utf8]{inputenc} 
\usepackage[T1]{fontenc}    
\usepackage{hyperref}       
\usepackage{graphicx} 
\usepackage{subfigure}

\usepackage{algorithm}
\usepackage{algorithmic}
\usepackage{amsfonts}
\usepackage{amsmath}
\usepackage{amssymb}
\usepackage[nohyperlinks,nolist]{acronym}
\usepackage{xparse}
\usepackage{wrapfig}

\usepackage{tikz}
\usepackage{array,multirow,multicol,booktabs,ctable}
\usetikzlibrary{arrows}
\usetikzlibrary{positioning}
\usepackage{hyperref}
\usepackage{wrapfig}

\graphicspath{{plots/}}

\begin{acronym}
    \acro{RNN}[RNN]{recurrent neural network}
\end{acronym}

\author{
  Tim Januschowski
  Jan Gasthaus
  Yuyang Wang
  David Salinas
  Valentin Flunkert
  Michael Bohlke-Schneider
  Laurent Callot\\
  Amazon Development Center\\
  Germany\\
  \texttt{<{tim.januschowski}@zalando.de>}
}
\begin{document}

\begin{frontmatter}

\title{Criteria for Classifying Forecasting Methods}

\author{}

\address{}


\end{frontmatter}

\begin{abstract}
Classifying forecasting methods as being either of a ``machine learning'' 
or ``statistical'' nature has become commonplace in parts of the forecasting literature 
and community, as exemplified by the M4 competition and the conclusion drawn by the organizers.
We argue that this distinction does not stem from fundamental differences in the methods 
assigned to either class. Instead, this distinction is probably of a tribal nature, 
which limits the insights into the appropriateness and effectiveness of different forecasting methods.
We provide alternative characteristics of forecasting methods which, in our view, 
allow to draw meaningful conclusions.  Further, we discuss areas of forecasting which could 
benefit most from cross-pollination between the ML and the statistics communities.
\end{abstract}

\maketitle

\section{Introduction}
This article discusses the spectrum of ``statistical'' and ``machine learning'' (ML) methods,
and the boundaries and intersections between them \emph{in the context of forecasting}.%
\footnote{We use quotes around ``statistical'' and ``ML'' here to emphasize
that, in our view, these are labels given to certain sets of techniques by some authors, not adjectives describing attributes of the methods.}
We argue that using the names ``ML'' and ``statistical'' to denote certain groups of techniques is unfortunate at
best, as they imply a more profound, qualitative distinction than the one they are
actually used to denote in practice.%
\footnote{To be concrete, the extensive, operational definition of ``statistical methods'' as used e.g.\ in \citep{makridakisM4, makridakisM4concl, Makridakis18} seems to be ``variants of exponential smoothing and ARIMA methods'', while ``ML methods'' is either used as an umbrella term to denote everything else or as a synonym for neural networks 
and random forests.}
In fact, implying a distinction using these general, connotation-laden terms can be misleading, as they incentivize 
sweeping statements such as ``[...] the accuracy of ML models is below that of statistical ones [...]'' \citep{Makridakis18}.

We argue that no meaningful grouping congruent with the original meaning of these terms can be
established, and that distinctions along other dimensions are more useful for
drawing valid conclusions in practice. We discuss how these dimensions allow to differentiate
the plethora of forecasting methods and contrast 
how they are approached by the statistics and ML \emph{communities}.
In the process, we attempt to clear up and correct certain misconceptions, 
such as \emph{ML methods do not handle uncertainty}~\citep{Makridakis18}.
	
While we make general claims in this paper, our aim is to focus the discussion on forecasting, 
not to contrast statistics and ML in general. The relationship between the fields of statistics 
and ML is complicated, partially due to the clash of ``two cultures,'' as put forth by~\citet{Breiman01statisticalmodeling}.
We are not oblivious to the observation that there are (partially) distinct ML and statistics \emph{communities} in forecasting.
It is our firm belief that these communities have much to learn from each other.\footnote{And indeed all
other communities that consider forecasting problems such as Econometrics, Data Science, and Systems, just to name a few.} We argue that it is more constructive to seek common ground than it is to introduce artificial boundaries.

\section{Connotations of the Terms ``Statistics'' and ``ML''}

Several authors have attempted to draw clear demarcation lines between statistics and ML.\footnote{Or try to establish clear distinctions between the these two and one or more of the following: Artificial Intelligence, Data Mining, Econometrics, Data Science, Pattern Matching, Predictive Analytics. See e.g.\ 
\url{https://towardsdatascience.com/no-machine-learning-is-not-just-glorified-statistics-26d3952234e3}\\
\url{http://andrewgelman.com/2008/12/machine_learnin/}\\
\url{http://brenocon.com/blog/2008/12/statistics-vs-machine-learning-fight/}\\
\url{https://www.svds.com/machine-learning-vs-statistics/}
} These attempts can be thought-provoking, but fall short of establishing clear distinctions and are mostly of limited scientific value.
Breiman~\cite{Breiman01statisticalmodeling} is an exception, as he proposes a distinction based on scientific culture rather than on a classification of methods. He distinguishes between a culture primarily concerned with in-sample fit (which he refers to as data modeling, explanation, or classical statistics) and one which is primarily concerned with out-of-sample fit (algorithmic modeling, prediction, or ML).

Breiman's article identifies a number of dimensions which tend to be used to classify methods as belonging to ML or statistics, such as: theoretical guarantees (statistics) vs.\ practical performance (ML) and mathematical elegance (statistics) vs.\ computational feasibility (ML). However, neither of these dimensions aligns with how these terms are used to distinguish forecasting methods.

Forecasting, as a field and discipline, draws from the qualities associated with
both terms ``statistics'' and ``ML'', making use of methods derived
 from formal probabilistic theory with statistical guarantees~\cite{durbin2012time,hyndman2008} 
 as well as purely algorithmic approaches such as Croston's method~\cite{Croston1972, Hyndman05}.
Forecasting is at its core an extrapolation
problem\footnote{There are of course in-sample fit concerns such 
as~\cite{KOLASSA2011238}, but these are not the main focus of
the discipline.}, where the performance of a model is evaluated using out-of-sample accuracy measures~\cite{Gneiting2007strictly,hyndman2006accuracy,kolassa07MAD} or tests~\cite{diebold2002comparing} rather than in-sample metrics.
Finally, we remark that forecasting has long embraced what~\citet{fiftyYears} identifies as the secret sauce of ML: focus on predictive modelling and the Common Task Framework (CTF).\footnote{\url{https://www.simonsfoundation.org/lecture/reproducible-research-and-the-common-task-method/}} An instance of CTF consists of a publicly available 
benchmark data set, a number of competitors working on the same (predictive) task and an objective referee. 
Through the M competitions~\cite{makridakisM4,m3,m2,m1}, forecasting has had a long lasting tradition in CTF.

In Section~\ref{sec:objective_dimensions} and \ref{sec:subjective_dimensions}, we discuss several 
dimensions along which forecasting methods can be  classified. Table~\ref{tab:dimensions} provides 
an overview. The dimensions we provide are by no means meant to be comprehensive, but should
be regarded as a starting point for discussion. Notable omissions are frequentist vs.\ Bayesian approaches,  
multivariate vs.\ univariate models or the assumptions underlying the models (such as Gaussianity or IID properties).\footnote{
We focus on dimensions which capture parts of the tension between ``statistical'' and ``ML'' forecasting methods. 
Other classifications (e.g., \citep{classification86,HBR71}) have provided further useful dimensions in more general contexts.
}

 The dimensions in Section~\ref{sec:objective_dimensions} are mathematical properties of the models, which is why we refer to them as \emph{objective} dimensions. In Section~\ref{sec:subjective_dimensions} we consider a set of dimensions that are of a methodological, or indeed cultural nature in the sense of \cite{Breiman01statisticalmodeling}, and we refer to them as \emph{subjective} dimensions. 
 The dividing line between objective and subjective dimensions is blurry, and we mainly provide this distinction in the hope of adding some additional structure to the exposition.
Section \ref{sec:oppo} synthesizes the discussion by identifying areas where we believe the interaction of both communities could be most fruitful.

\begin{table*}[h!]
\begin{center}
\begin{tabular}{cllr}
\toprule
Category & Dimension & Section \\
\midrule
\multirow{5}{*}{Objective}
& Global vs. Local Methods & \ref{sec:gl}\\
& Probabilistic vs. Point Forecasts& \ref{sec:unertainty}\\
& Computational Complexity & \ref{sec:comp}\\
& Linearity \& Convexity & \ref{sec:lin_cvx}\\
\midrule
\multirow{5}{*}{Subjective}
& Data-driven vs. Model-driven &  \ref{sec:data}\\
& Ensemble vs. Single Models &  \ref{sec:ensemble}\\
& Discriminative vs. Generative &  \ref{sec:discriminative}\\
& Statistical Guarantees &  \ref{sec:theory}\\
& Explanatory/Interpretable vs. Predictive &  \ref{sec:explanatory}\\
\bottomrule
\end{tabular}
\caption{Summary of all dimensions discussed.}\label{tab:dimensions}\end{center}
\end{table*}

\section{\emph{Objective} Dimensions for Classifying Forecasting Methods}\label{sec:objective_dimensions}

This section considers a set of \emph{objective} dimensions along which forecasting methods can be classified. We contrast how the statistics and ML communities tend to address them and highlight commonalities and complementarities.

\subsection{Global and Local Methods}
\label{sec:gl}
Consider a forecasting problem in which we aim at predicting a large number of similar time series, for instance, predicting demand for similar products. To tackle such problems, we can distinguish between two extremes:
methods that estimate model parameters independently for each time series (here referred to as \emph{local} methods) and methods that estimate model parameters jointly from all available time series (here referred to as \emph{global} methods)~\cite{januschowski18}.\footnote{Note that historically, local methods have also referred to methods such 
as local linear methods e.g.,~\cite{Bottou1992,weigend93} and~\cite{Bontempi2013} and references therein. This is an unfortunate overlap in terminology, but the difference should be clear from the context.}

There are hybrids between these extremes~\cite{geweke1977dynamic, stock2002macroeconomic, taieb17icml,deepFactor,deepFactor19}, 
where a part of the parameters are estimated globally and a part locally.

Note that this distinction is about how the parameters of the model are estimated, and
does not imply a particular (in-)dependence structure between the time series. In particular,
one can estimate a global model and still assume independence between forecasts for different time series (e.g.\ for reasons of computational efficiency). As such, this distinction is complementary to the distinction between univariate and multivariate forecasting methods, where multivariate typically implies an explicitly modeled dependency structure between the time series 
(e.g.,\ explicitly modelling the co-variance structure).

While many ``statistical'' methods are local methods,
global methods have long been used in both the statistics and ML communities. One key difference (and source of some confusion) is that models such as NNs can be used both in global and local settings without modifications. Recently, NNs have been used primarily as global models~\cite{flunkert2017deepar,kari2017,laptev2017}, thereby surpassing earlier, mixed results where NNs where mainly used as local models~\cite{zhang1998forecasting}.
In local settings, only models with a small number of free parameters can typically be fitted reliably as training data is limited. On the other hand, global NN models with millions of parameters can be successfully trained as long as a large-enough training set of related time series is available. 

\subsection{Uncertainty Quantification and Distributional Forecasts}
\label{sec:unertainty}
Forecasting methods can be classified into probabilistic and point forecasting methods. 
While point forecasts just provide a single, best prediction (relative to some error metric)~\cite{Gneiting09,Gneiting11}, probabilistic forecasting methods quantify the \emph{predictive uncertainty}, allowing this uncertainty to be taken account when making decisions based on the forecast.
Uncertainty in forecasts (predictive uncertainty) has multiple sources, including: (a) uncertainty about the model's parameter values (e.g.\ the slope of a trend); (b) uncertainty about the model's structure (e.g.\ whether to include in linear or quadratic trend); (c) residual variation, i.e.\ randomness not explained by the model (either due to inherent randomness of the underlying phenomenon, or due to simplifying assumptions made by the model); (d) uncertain model input data (e.g., imprecisely measured input data used for training or the need to forecast causal drivers for prediction); and, (e) uncertainty about the stability of the model or model drift, where the causal drivers of the forecast or their relation to the target may change over time. Residual variation is often treated explicitly in forecasting models in the form of explicit noise terms. Parameter and model structure uncertainty are sometimes also taken into account, by either Bayesian (e.g.\ averaging model predictions with respect to the posterior distribution over parameters and/or models), or frequentist approaches (e.g.\ model ensembles and bootstrap sampling).

For a time series $z_1, z_2, \ldots, z_{T}$, the predictive uncertainty is fully characterized by the 
predictive distribution 
\begin{equation}
P(z_{T+1}, z_{T+2}, \ldots, z_{T+h}|z_1, z_2, \ldots, z_T)\,, \label{eq:pred_dist}
\end{equation} 
but probabilistic forecasting methods differ in how they allow the user to query this (typically intractable) joint distribution. Typical ways include pointwise predictive intervals (i.e.\ access to quantiles of the marginal predictive distribution for each forecast horizon), as well as Monte Carlo sample paths from the joint predictive distribution. Another, commonly-used option is to assume a parametric form of the distribution (e.g.,~\eqref{eq:pred_dist} is a negative binomial distribution) and return the parameters.

Marking probabilistic forecasting methods as ``statistical'' and point forecasting methods as ``ML'' is implied in~\cite{Makridakis18}\footnote{
The authors write that ``at present, the issue of uncertainty has not been included in the research agenda of the
ML field, leaving a huge vacuum that must be filled as estimating the uncertainty in
future predictions is as important as the forecasts themselves.''}. 
However, there is ample evidence that probabilistic modeling and handling various forms of uncertainty has been at the heart of much of ML for at least the last decade, exemplified by most introductory text books taking a probabilistic perspective \citep{bishop2007, Barber12, Koller2009, Murphy2012}. 
Modern ML methods handle uncertainty, e.g.\ by postulating generative models whose parameters are estimated through maximum likelihood estimation \citep{flunkert2017deepar}, or by estimating the quantile function directly~\cite{kari2017,quantileRnns} similarly to the approach of \cite{koenker_2005} in econometrics.
Estimating model and parameter uncertainty using Bayesian approaches is a well developed area in the ML literature as well; Barber~\cite{Barber12} provides an
introductory overview and for recent contributions see~\cite{vae2013,blundell2015weight, Gal2016Uncertainty}. 

Furthermore, we note that the results of the M4 competition have shown that prediction intervals obtained by methods from the ML community have proven to be highly accurate~\cite{makridakisM4} even though they may lack a theoretical underpinning, thereby directly contradicting~\cite{Makridakis18}.

\subsection{Computational Complexity and Costs}
\label{sec:comp}
As forecasting is becoming more wide-spread and is applied to increasingly large data sets,
it is natural to consider a notion of computational cost in addition to predictive accuracy when comparing forecasting methods, and to consider the trade-off. Depending on the application, the size of the downstream impact through improvements in forecast accuracy by a computationally more demanding method may be outweighed by the increased computational cost. Different applications may place constraints on the maximum time that can be spent on generating a forecast for a single time series (e.g.\ in (near) real-time settings) or impose a limit on the time available to produce 
forecasts for the entire body of time series. If there's an application-specific absolute time limit, parallelizability becomes an important factor. So, for a data set consisting of many items that need forecasting, it may be both interesting to measure
the maximum time to forecast a single item (this would be the bottleneck in a highly-parallelizable scenario) as well as measuring the cumulative time to produce all forecasts.

For forecasting tasks where forecasts need be produced periodically (e.g.\ retail demand forecasts
produced daily or weekly as input to an automatic ordering system), the computational cost of a forecasting method can be broken down into three components: (1) the computational cost of the experimental phase of finding an adequate model and model hyperparameters, (2) the computational cost of \emph{training} the forecasting model, and (3) the computational cost of producing forecasts from a trained model (this is commonly referred to as \emph{inference} 
in the deep learning literature). Note that in popular implementations of local models, (2) and (3) are conflated
as the parameters are re-estimated for each forecast, while for global models, (2) and (3) are separate steps. This allows the cost of training (2) to be amortized over multiple forecasts (3), e.g.\ by re-training a model only once per month but using it daily to update the forecasts.
There are some further subtleties to consider here, as both local and global methods may 
benefit from warm-starting using previously obtained parameters/states, and steps (2) and (3) benefit from parallelization in different ways. Prediction is typically embarrasingly parallel (i.e., parallelizable over the data), while training only achieves sub-linear speedup (primarily for global models; training for local models  
is embarrasingly parallel). 

While some deep learning-based forecasting methods may require lengthy training steps (though even these are nowadays on the order of hours, not days or weeks, on data sets with millions of time series), they are typically comparable in running time to local ``statistical'' methods when making forecasts.
For example, one of the most widely used ``statistical'' forecasting algorithm implementations is the \texttt{ets} from~\cite{hyndman2008R}
due to its robustness and efficiency.\footnote{In particular, with the default settings \texttt{model = "ZZZ"} which uses model selection. We use version 8.5 of the Forecast package.} This robustness is achieved by running several models and keeping
the one with the best Akaike Information Criterion.
On a large weekly retail demand dataset~\citep{seeger2016}, \texttt{ets} takes around 30 ms to produce a forecast for a single time series (i.e.\ a full model selection/training/prediction cycle).
The recurrent neural network model (RNN) of \citet{flunkert2017deepar} requires about one hour for training and can then generate forecasts in around 100 ms per time series on a commodity laptop with no GPU. While \texttt{ets} is indeed faster, it is noteworthy that the time differential between the two methods is smaller than could be expect given the difference in model complexity. In addition, training time can be amortized when forecasts are made periodically without re-training. In a production setting, we have seen that more than 80\% of the compute costs are incurred during inference and only 20\% during training.

This is very much in contrast with the experimental stage of model building where most of the time is spend on training, as that is what is needed to improve models.

The M4 competition and in particular~\cite{makridakisM4} has taken a first step
to take computation time into account.  

In addition to computational costs, other costs may need to be considered as well: A forecasting method that requires supervision and tuning by an expert human operator may ultimately be more expensive than a computationally demanding but fully automated method.

\subsection{Linearity \& Convexity}
\label{sec:lin_cvx}
Classifying methods into linear and non-linear classes is common place
in most mathematical disciplines, as it is typically easier to establish formal results for the class of linear methods than it is for non-linear methods.

In the forecasting setting, there are multiple aspects of a model which can be (non-)linear, e.g.\ the forecast can be a linear function of the past observations (as in linear autoregressive models), of the parameters, of the covariates (as in linear regression), or of time (i.e.\ models with linear trend); the underlying dynamical system can have linear dynamics; or the underlying optimization problem can have a linear objective function and/or constraints.

While some ``statistical'' methods are linear in one or more of the above senses, and many ``ML'' methods are non-linear, conflating linear with ``statistical'' and non-linear with ``ML'' methods as in \citep{Makridakis18} is an overgeneralization, as there exist ``statistical'' methods that are not linear in any meaningful way (e.g.\ exponential smoothing with damped or multiplicative trend), and there also exist ``ML'' methods that are linear (e.g.\ linear support vector regression \citep{smola2004tutorial} or matrix factorization \citep{NIPS2016:6160}).

Similar to linearity, convexity of the underlying objective function of a learning/forecasting algorithm is an attribute that is often considered, as convex optimization problems are generally tractable. Postulating a convex loss function is a straight-forward way of ensuring that a practical
optimization procedure (e.g.\ gradient descent) can identify a global optimum instead of
just a local one. A non-convex loss function on the other hand makes identifying a global
optimum practically impossible in most cases, making it challenging to establish theoretical
guarantees for such methods. However, the statistics and ML communities have found non-convex models to be effective for many applications~\cite{sockeye,wavenet,Goodfellow-et-al-2016}, including forecasting.

Forecasting methods widely used in the statistics community---such as variants of the generalized
autoregressive conditional heteroskedasticity model---require solving non-convex optimization problems.
The parameter identification problem, which occurs when several parametrizations of a model
are observationally equivalent, is a core issue in the field of econometrics~\cite{fisher1966identification}
and stems from loss functions that have more than one global optimum and are therefore not strictly-convex.

In the field of ML, the regain in popularity of NNs triggered much discussion on the topic
of (non-)convex loss functions. The skepticism towards models with non-convex loss functions
(which includes almost all NN-based models) was overcome by their outstanding predictive
performance in practice, helped by refinements in (mainly stochastic) gradient descent methods.
Recent research in the deep learing community has to some extent validated these practical results by
showing that for sufficintly large NNs almost all local minima are very similar to the global minimum 
(see e.g.~\cite{choromanska2015loss}).

\section{\emph{Subjective} Dimensions For Classifying Forecasting Methods}\label{sec:subjective_dimensions}

In this section we turn to subjective dimensions, dimensions of a methodological or cultural nature.
While they are naturally less well-defined than objective dimensions, considering them is helpful since we believe
that it is along these dimensions that the principal differences between the ``ML'' and ``statistics'' approaches to
forecasting can be found.

\subsection{Data-driven and Model-Driven Methods}\label{sec:data}

Methods that are generally considered as belonging to the realm of ML, such as NNs or random forest~\cite{Breiman01},
are mostly data driven, in the sense that such methods memorize patterns effectively and do not make
strong structural assumptions (such as ``the trend has to be linear'' or ``the change from one point to the
next must be smooth''). Data-driven models are flexible at the cost of being data-hungry, in the sense that they typically have a large number of parameters and require a sufficient amount of data to tune those parameters.

\begin{figure}[!ht]
\includegraphics[width=\textwidth]{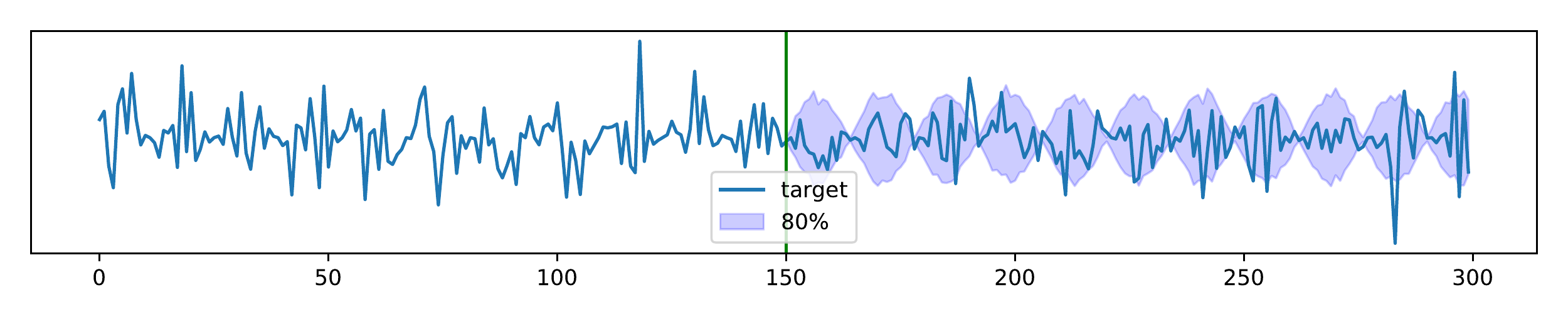}
\caption{Plot of the 80\% prediction interval for the 150 units following the last point seen in a training set consisting of observations of an heteroskedastic cyclic noise $z_t = \varepsilon_t ~\text{sin}(t)$ with $\varepsilon_t \sim \mathcal{N}(0, 1)$. \label{fig:sinusoid}}
\end{figure}

In Figure \ref{fig:sinusoid}, we show an example where the recurrent neural network model from \cite{flunkert2017deepar} is trained to predict a heteroskedastic noise with an oscillating variance amplitude\footnote{More precisely, this artificial dataset consists of one time series $z_t = \varepsilon_t ~\text{sin}(t)$ with $\varepsilon_t \sim \mathcal{N}(0, 1)$. We train with $N$=1M observations with the values $z_t$ with $t<N$ and plot the 80\% prediction interval forecasted for the 150 times units following $t = N$, e.g. the last point seen in the training. Note that the frequency of the noise is inferred only from the data.}. From this example, one can see that a complex non-linear time series pattern can be retrieved only from the data. Clearly, being able to infer complex behavior only from data comes with the downside risk of over-fitting. Regularization schemes such as Dropout~\cite{dropoutBaldi} can help alleviate
 such effets. While it has been shown for instance that state of the art architectures can fit very large corpora of images almost perfectly with random labels \cite{Zhang2016}, the case of time series may be a bit different. Indeed, recurrent neural networks can not memorize more than 5 bit of information per parameter \cite{Collins}. In this context, determining the amount of data (or regularization) required to avoid over-fitting in the case of forecasting remains an open problem.

At the other end of the spectrum are models generally considered as belonging to the realm of statistics, such as ARIMA models and GLMs, that are parsimoniously parameterized and therefore need little data to be
accurately fitted. Model-driven approaches are more rigid in the sense that they can only model a
limited set of patterns defined by the assumptions made when specifying the model.
If these assumptions are correct, these models are very data-efficient as they need little
data for their parameters to be accurately estimated. Counter-intuitively, systematically misspecified models 
may still lead to better predictions than correctly specified models~\cite{kolassa2016}.

In our experience, data-driven models tend to be a good choice when used as global
models employed for operational forecasting problems~\cite{janusch18}, where they
are trained on a large number of time series. For these problems, they are able to extract complex
patterns with little intervention from the researcher. Model-driven approaches
on the other hand often need careful feature engineering and specification to be
able to adequately capture the regularities in the data, and can be a very efficient
choice when dealing with a sufficiently small number of series for which the researcher
is able to specify an appropriate model.

Smyl's winning solution to the M4 competition \citep{smyl} shows that models in which data-driven and model-driven methods interact can perform extremely well, and more research in understanding
when and how such beneficial interactions occur is needed. For example, in Smyl's approach, a natural next experiment could be an ablation study to tease out the
effect of the data-driven and the model-driven
parts of the solution on the overall accuracy.

\begin{table*}[h!]
\begin{center}
\begin{tabular}{lll}
\toprule
Metric & DeepAR & Smyl \\

\midrule
sMAPE & 0.1192 & 0.1137 \\
MASE & 1.500 & 1.54\\{}
owa & 0.837 & 0.821\\
MSIS & 12.07 & 12.23\\
\bottomrule
\end{tabular}
\caption{Comparison between a pure RNN and a hybrid RNN method for the M4 dataset.}
\label{tab:DeepARM4}{}
\end{center}
\end{table*}

Table~\ref{tab:DeepARM4} contains the results of a data-driven method~\cite{flunkert2017deepar}
as implemented in~\cite{januschowski18} with default hyper-parameters. The method uses a recurrent
NN but no exponential smoothing equation. Its accuracy is close to the winning solution
in the M4 competition.

\subsection{Ensembles, Model Combinations, and Single Models} \label{sec:ensemble}

Section~\ref{sec:lin_cvx} introduced a dimension which allows to distinguish
simple from complex models. Another
dimension along which we can differentiate between simple and complex models is to
single out ensemble methods as exemplified in the M4 competition.

The M4 competition has a single entry for both combination and single models, but does mark a
method as a \emph{combination} method (Table 4 in~\cite{makridakisM4}). The extremes on this distinction
are clear. One the one extreme are methods such as~\cite{FFORMA} where a gradient-boosted tree~\cite{xgboots}
is used to combine multiple models. On the other extreme is a fully specified model with fixed hyper-parameters
where only the primary parameters are estimated. 

In between both extremes, there's considerable ambiguity. Consider a method that automatically selects the best model structure
in the class of ARIMA models (e.g.\ \texttt{auto.arima}). This method does result in a single model being used to produce a forecast,
but to select this particular model multiple models were estimated and evaluated on some goodness of
fit or out-of-sample accuracy criterion. The same ambiguity exists, for example, for GLMs that rely on automated feature selection techniques or for any model that makes use of hyper-parameter optimization procedures~\cite{pmlr-v70-jenatton17a}, mixtures of experts~\cite{bishop2007}, or drop-out regularized NNs~\cite{dropoutBaldi}.

To us, it remains unclear where on the continuum of single models vs. combination models the above examples 
would fall. Despite its ambiguity, the value of this dimension comes from its separation of concerns. It allows to isolate the assessment of forecasting models from combination techniques and it allows for studies comparing end-to-end learnt approaches (such as NNs) against staged approaches and ensembles.

\subsection{Discriminative \& Generative Models}\label{sec:discriminative}

We can in general distinguish between discriminative and generative models in statistics and ML and hence,
also in forecasting. For a time series $z_1, z_2, \ldots, z_{T}$ which we want to forecast,
a generative forecasting model aims at modelling $P(z_1, z_2, \ldots, z_{T}, \ldots, z_{T+h})$.

In contrast, a discriminative forecasting method directly models this predictive distribution
\[
P(z_{T+1}, z_{T+2}, \ldots z_{T+h}| z_{1},\ldots, z_{T}).
\]

Often, generative models decompose $P(z_1, z_2, \ldots, z_{T}, \ldots, z_{T+h})$ into telescoping conditional distributions 
\[
P(z_1, z_2, \ldots, z_{T+h}) = P(z_1)P(z_2|z_1)P(z_3|z_1, z_2)\cdots \cdot P(z_{T+h}|z_{1}, \ldots, z_{T+h-1}).
\]
Using Bayes rule, we can then obtain the predictive distribution
for making forecasts for periods $T+1,\ldots, T + h$, conditioned on the past
\[
P(z_{T+1}, z_{T+2}, \ldots z_{T+h}| z_{1}, z_2, \ldots, z_{T}).
\]

 Note that while this distinction seems rigorous at first for forecasting models, there are models
 that can be used both as discriminative and as generative models, hence we decided to place this
 distinction in the subjective category.

Different NN architectures for forecasting can be compared along this dimension. ~\citet{sutskever2014} provide a NN for discriminative, neural sequence models while concurrent
research in the ML community has also provided generative models based on NNs (e.g.,
~\cite{fraccaro2016sequential,rangapuram2018,krishnan2015deep}).

Understanding the differences between discriminative and generative models can trigger relevant research.
For example, discriminative
models may be more accurate in certain settings at the expense of model limitations (for example, models
such as~\citep{sutskever2014} must be re-trained if $h$ changes). 
Finally, and to the best of our knowledge, there are no general theoretical results establishing the superiority of one approach over the other in a given setting, with the exception of~\cite{ng02} for logistic regression. 

\subsection{Statistical Guarantees \& Bounds}\label{sec:theory}

A reader familiar with the scientific literature published in statistical journals and ML venues
will have been struck by the structural differences in the articles.

An article published in a statistics journal\footnote{We invite the reader to refer to articles
from statistics journals cited in this paper.} proposing some new method (model, estimator, or test)
typically starts by stating a series of assumptions on a data generating process.
It then establishes the properties of the proposed method, often in the form of asymptotic results
or finite sample bounds. The authors continue by reporting the results of an empirical study (typically small-scale and partially on synthetic data)
aimed at validating the theoretical results and comparing to existing benchmarks,
occasionally followed by a ``real-world'' application demonstrating the practical relevance of the proposed method.

A paper published in an ML venue\footnote{We invite the reader to refer to articles from ML
journals or conferences cited in this paper.} proposing a new procedure generally starts by
describing that procedure and motivating its use. The authors then seek to contextualize
the proposal within the existing literature before reporting results of experiments
using the proposed method on several standard datasets and comparing it to benchmark procedures.

The description of articles in ML and statistics outlets in the two paragraphs above is
somewhat caricatural and certainly is an over-generalization (there are application-focussed statistic papers, just as there are purely theoretical ML papers), but it does reflect a
substantial methodological difference between both research communities. This methodological
difference is, in our view, one of the main source of tension between the statistics and ML
communities. While the statistics approach tends to favor providing theoretical guarantees
based on assumptions that are not always testable (and routinely violated in practice), the complex models
used in the ML community are usually not amenable to full theoretical analysis and the
ML community therefore relies heavily on validation through empirical evaluation of
out-of-sample performance. These purely experimental results can be hard to verify since the outcome can hinge on
details of the experimental setup. The reproducibility concern has been recognized in the ML community 
(e.g., major ML outlets such as KDD and NeurIPS now ``expect'' source code). Even if the experiments are reproducible, 
the results are not necessarily generalizable to
other datasets and problems. Furthermore, an over-reliance on standardized benchmarks may result in overfitting on these benchmarks. 

\subsection{Explanatory Modeling \& Interpretability} \label{sec:explanatory}
Forecasting is primarily concerned with \emph{predictive modelling} rather than
\emph{explanatory modelling} \citep{shmueli2010explain}. Still,
interpretability, e.g.\ defined as ``the degree to which an observer can
understand the cause of a decision'' \citep{miller2017explanation}, of
forecasting models often high on the list of desiderata despite the fuzzyness of the concept.
``ML'' methods are often seen as a black boxes, and therefore
inherently not amenable to interpretation.
Indeed, the complex, non-linear mappings implemented by NNs, for example,,
make interpretation challenging. This is an active research topic~\cite{lime, MonSamMue18,cleverHans19}.

In contrast, because of their parsimonious parametrization linear models are often viewed as being straightforward to interpret. 
This simplicity is deceptive though.
Consider adding to an estimated model a new covariate that is not orthogonal with those already included in the model.
This will modify the partial correlations estimated by the model, including possibly the sign of the associated parameters.
As a consequence, the interpretation of the model will change.  
The amount of structural insight on the data generating process that can be gained from looking at the partial correlations estimated by a linear model is limited and contingent.

In the particular case of forecasting, non-linear transformations of the raw input data
(which are at the core of NNs) and feature engineering are often crucial for
achieving high accuracy and the success of many classical methods. These transformations often hinge
on complex (and sometimes manual) data pipelines that transform data, e.g.\ by de-seasonalization,
de-trending, and other complex pre-processing steps
(see e.g.,~\cite{hyndman2017forecasting, seasonalityIndices12}). Even if we assume the
core model to be interpretable, it is only so in
light of its direct inputs which have been subject to difficult to interpret
transformations~\cite{bose2017probabilistic}. A classic example of this are cold start (e.g.\ new product)
forecasts~\cite{np2018}, where many approaches require determining the similarity
between items (via judgmental approaches, clustering or embeddings in modern ML parlay).
Most of these are highly non-interpretable.

\paragraph{Causality}
Interpreting partial correlations does not help understand ``the cause of a decision''
 \citep{miller2017explanation}. Causal modeling is an active research topic in the statistics
 literature~\cite{imbens2015causal, granger2001essays, pearl2009causal} as well as in
 the ML literature~\cite{pearl2009causality, peters2017elements}.
 Research on causality at the intersection of traditional ML and statistics methods
 provides an example of fruitful cross-pollination between approaches traditionally
 classified as statistical (or econometric) and ML, see for example
 \cite{athey2015machine, chernozhukov2016double, wager2017estimation},
 but this topic is beyond the scope of the present paper.

\section{Opportunities \& Challenges}\label{sec:oppo}
The previous sections aimed at showing that the methods developed by the statistics and ML communities share a lot of concepts but that these communities have developed divergent cultures and vocabularies, and emphasized different topics in their research agendas.
We believe that they can learn and benefit from each others' strengths. The conclusion of this paper discusses what we view as the most promising opportunities.

\paragraph{Software Frameworks}
ML frameworks have reached a high degree of sophistication over the last years and
extended their reach beyond pure deep learning. These frameworks allow for a quick turn-around in model development. This is enabled by many components coming together in these frameworks: fast computational primitives on tensors, auto-differentiation, implementations of sophisticated optimization algorithms, probabilistic
tools such as distributions and sampling algorithms, allowing e.g.\ the implementation of MCMC frameworks~\cite{bingham2018pyro,anonymous2019modular}.

MXNet~\cite{mxnet} for example even offers an R front-end, although most deep learning frameworks focus on Python. Many application-specific toolkits exist (e.g., for machine translation~\cite{sockeye}, Gaussian Processes~\cite{mxfusion}, computer vision, natural language processing\footnote{
	\url{https://kdd18.mxnet.io/}
	\url{https://github.com/dmlc/gluon-cv}
}) and it is only a question of time until one for forecasting will be available.\footnote{https://eng.uber.com/m4-forecasting-competition/}

The convergence on an open-source ecosystem for forecasting based on these deep learning frameworks will allow for greater reproducibility
of experiments, better model exchange, and help popularize these methods, much like the R \texttt{forecast} package has stimulated research.

\paragraph{Empirical Rigor}
The M-competitions are examples for the rigor with which the forecasting community approaches
empirical experiments. Adopting this rigor in the empirical evaluations for forecasting methods
which is also present in journal such as the IJF would be a large step forward for the ML community.
~\citet{Makridakis18} correctly points this out. The ML community working on forecasting should compare against
the state of the art in the forecasting community and commonly accepted baselines such as the R \texttt{forecast} package~\cite{hyndman2008R}, e.g.,~\cite{kari2017,mdn2018}. Conversely, a number of forecasting competitions are available on Kaggle.\footnote{
For example:
\url{https://www.kaggle.com/c/favorita-grocery-sales-forecasting/data}
\url{https://www.kaggle.com/c/walmart-sales-forecasting/data}
\url{https://www.kaggle.com/c/rossmann-store-sales/data}
\url{https://www.kaggle.com/c/recruit-restaurant-visitor-forecasting}
\url{https://www.kaggle.com/c/global-energy-forecasting-competition-2012-load-forecasting/data}
} Conclusions from these competition such as the success of ensemble and data-driven methods have
foreshadowed the results of the M4 competition and could be taken more seriously in the academic literature.

It is important to note that we should limit the conclusions based on empirical evidence
to the data sets on which the methods are evaluated. The forecasting landscape is rich and
contains many different forecasting tasks and data sets.
If the benchmark dataset only covers certain domains within forecasting or focuses on a
single forecasting task (e.g., only on cold-start forecasting), it is important to clearly
define the intent of the experiment and to restrict the conclusions drawn from the
results of an experiment to this context. For example, we expect data-driven methods
to perform well on operational forecasting
problems and model-based methods on strategic forecasting problems~\cite{janusch18}.

\paragraph{Practical considerations}
Success in forecasting applications depends on many details that are often brushed under the carpet by
the different communities to varying degrees. We'll provide a few examples in the following. Carefully
crafting the training schemes of forecasting models, such as~\cite{flunkert2017deepar,laptev2017}
is an important predictive performance boost. Designing co-variates and making them available
in software packages,\footnote{
For example	\url{https://tsfresh.readthedocs.io}
	\url{https://github.com/thiyangt/seer}
} 
 handling them appropriately, in particular
in the context of global models~\cite{flunkert2017deepar}, assembling pipelines of
models~\cite{bose2017probabilistic,seasonalityIndices12}, modern hyper-parameter optimization
 techniques (see e.g.,~\citep{pmlr-v70-jenatton17a}), model combination techniques (Section~\ref{sec:ensemble})
 or the re-use of already trained models and topics such as
transfer and meta-learning are a selection of the many areas where the communities can learn from each other.
The requirements of modern forecasting scenarios via high-dimensional, streaming or big data use cases (e.g., via internet of things applications) amplify the importance of such practical considerations.

\paragraph{Theory}

Section~\ref{sec:theory} describing the difference between papers published in statistics and in ML journals noted that theoretical guarantees are not considered necessary to a contribution in ML. This is an issue that is debated within the ML community with researchers pointing out that the lack of a theoretical explanation for the effectiveness of deep learning methods\footnote{See A.~Rahimi \emph{NIPS 2017 Test-of-Time Award} acceptance speech: https://www.youtube.com/watch?v=ORHFOnaEzPc} puts continued progress in the field at risk.

Other researchers in the ML community argue that the lack of theoretical foundations is not a sufficient reason to reject a method that has proven its effectiveness empirically. This view can be supported by considering the history of the Lasso estimator. The original article proposing the Lasso \cite{tibshirani1996regression}, published in a leading statistical journal, contains little theoretical results on the method but is mainly descriptive. After the Lasso proved to be an extremely useful estimator, a large body of research was dedicated to studying its properties in a wide range of settings, for times-series and forecasting see \cite{kock2015oracle, chan2014group, basu2015regularized} among many others.

NN methods have proven their usefulness across a large number of applications, but the theoretical understanding is lagging. The gap is beginning to be closed though, see e.g. \cite{2018arXiv180909953F} for NNs and \cite{athey2016generalized} for random forests. This is an area in which collaboration between the ML and statistics community could be beneficial to both.

\section{Conclusion}
\label{sec:conc}

While we believe that there is no added value in classifying methods according to
being an ML or a statistical method, we do think that making this distinction
points to a larger, important issue. The scientific communities working on forecasting, while working
on the same problem, do not nearly interact as much as they should.  The communities
working on the forecasting problem have contributed relevant results and we therefore encourage all readers of this
article to step outside their comfort zone.

We note that there are encouraging steps. For example, current research
is concerned with the combination of probabilistic models with
NNs~\cite{krishnan2015deep,januschowski18,fraccaro2016sequential,krishnan2017structured,fraccaro2017disentangled,rangapuram2018}.
There are many more examples of such hybrids from methods from the ML and the statistical
community where we can use gradient-boosted trees to select classical methods (see e.g., \cite{FFORMA}
and runner-ups from the M4 competition), using ML approaches to improve multi-step ahead forecasting~\cite{pmlr-v32-taieb14,Bontempi2013} or pipelines of models
which contain a combination of ML and statistical methods~\cite{bose2017probabilistic}. Similar
to~\citep{makridakisM4}, we expect much progress to come from this line of work and we look
forward to the M5 competition.

\section*{Acknowledgements} We thank Stephan Kolassa for inspiring discussions
on the topic.

\section*{References}

\bibliography{mybibfile}

\clearpage

\end{document}